\documentclass[letterpaper,10pt,conference]{ieeeconf} 

\IEEEoverridecommandlockouts                              

\usepackage{graphics} 
\usepackage{epsfig} 
\usepackage{mathptmx} 
\usepackage{times} 
\usepackage{amsmath} 
\usepackage{amssymb}  
\usepackage{xcolor}
\usepackage{algorithm2e}
\makeatletter
\renewcommand{\@algocf@capt@plain}{above}
\renewcommand{\algocf@caption@plain}{\box\algocf@capbox\vskip\AlCapSkip}%
\makeatother

\usepackage{balance}
\usepackage[utf8]{inputenc}
\usepackage[pdfauthor={Bonetto et al.},
            pdftitle={GRADE: Generating Realistic Animated Dynamic Environments},
            pdfsubject={Dynamic environments generation pipeline},
            pdfkeywords={slam,mapping,localization,learning,dynamic,animations,isaac,dynamic environments,dataset}]{hyperref}
\let\labelindent\relax
\usepackage{enumitem}
\usepackage{subcaption}
\usepackage{cuted}
\usepackage{multirow}
\hypersetup{
    colorlinks=true,
    linkcolor=blue,
    filecolor=magenta,      
    urlcolor=blue,
}

\usepackage[font=footnotesize,labelfont=bf]{caption}
\usepackage{tikz}

\usepackage[textsize=tiny]{todonotes}

\newcommand{\uproman}[1]{\uppercase\expandafter{\romannumeral#1}}

\usepackage{eso-pic}
\usepackage{soul}

\title{\LARGE\bf{Simulation of Dynamic Environments for SLAM}}

\author{Elia Bonetto$^{*,\dagger}$ ~\IEEEmembership{Student Member,~IEEE,} Chenghao Xu$^{\ddagger,*}$, and Aamir Ahmad$^{\dagger,*}$~\IEEEmembership{Senior Member,~IEEE}
    \thanks{$^*$Max Planck Institute for Intelligent Systems, Tübingen, Germany. {\tt\footnotesize {firstname.lastname}@tuebingen.mpg.de}}
    \thanks{$^\dagger$Institute of Flight Mechanics and Controls, University of Stuttgart, Stuttgart, Germany. {\tt\footnotesize {firstname.lastname}@ifr.uni-stuttgart.de}}
    \thanks{$^\ddagger$Faculty of Mechanical, Maritime and Materials Engineering, Department of Cognitive Robotics, Delft University of Technology, Delft, Netherlands.}%
	\thanks{The authors thank the International Max Planck Research School for Intelligent Systems (IMPRS-IS) for supporting Elia Bonetto.}
}

\begin{document}

\AddToShipoutPictureBG*{%
  \AtPageUpperLeft{%
    \setlength\unitlength{1in}%
    \hspace*{\dimexpr0.42\paperwidth\relax}
    \makebox(0,-0.75)[c]{\parbox{0.8\textwidth}{\textbf{2023 IEEE International Conference on Robotics and Automation (ICRA) }\\
    \textbf{Workshop on Active Methods in Autonomous Navigation, \href{https://robotics.pme.duth.gr/workshop_active2/?page_id=490}{URL}}\\
    \textbf{29 May - 2 June, 2023, London, UK}}}%
}}

	\maketitle
\begin{abstract}

Simulation engines are widely adopted in robotics. However, they lack either full simulation control, ROS integration, realistic physics, or photorealism. Recently, synthetic data generation and realistic rendering has advanced tasks like target tracking and human pose estimation. However, when focusing on vision applications, there is usually a lack of information like sensor measurements or time continuity. On the other hand, simulations for most robotics tasks are performed in (semi)static environments, with specific sensors and low visual fidelity. To solve this, we introduced in our previous work a fully customizable framework for generating realistic animated dynamic environments (GRADE)~\cite{GRADE}. We use GRADE to generate an indoor dynamic environment dataset and then compare multiple SLAM algorithms on different sequences. By doing that, we show how current research over-relies on known benchmarks, failing to generalize. Our tests with refined YOLO and Mask R-CNN models provide further evidence that additional research in dynamic SLAM is necessary. The code, results, and generated data are provided as open-source at~\url{https://eliabntt.github.io/grade-rr}.

\end{abstract}


	\section{INTRODUCTION}
\label{sec:intro}

Intelligent robots should be able to perceive and understand the world around them to be autonomous and able to interact with it. However, especially when addressing dynamic environments, it is not possible to experiment with them directly in the real world due to the inherent risk of damaging or hurting people and animals. Therefore, it is crucial to verify them beforehand in simulation. 

Gazebo~\cite{gazebo} is by far the most popular framework for simulating robots due to its simplicity, reliable physics engine, and tight integration with ROS~\cite{ros}. Nevertheless, it is not photorealistic, there is a limited variety of assets/worlds that can be loaded without considerable effort, and the simulation engine cannot be fully controlled. For these reasons, various alternatives emerged, e.g. TartanAir~\cite{tartanair}, AirSim~\cite{airsim}, AIHabitat~\cite{habitat19iccv}, BenchBot~\cite{benchbot}, and iGibson~\cite{igibson}. However, they all lack either full control of the simulation, ROS integration, realistic physics and appearance, or SIL/HIL capabilities. Additionally, some simulate environments with \textit{only} rigid objects~\cite{gazebo,habitat19iccv} or do not include dynamic assets since this would pose various challenges such as their placement, their management, and their generation. Then, relying on pre-recorded robotic datasets is typically non-trivial due to differences in the form factors of the robots (e.g. placement of the sensors, stiffness of the joints), in the sensor settings (e.g. focal length, FPS, IMU frequency) or in the noise models. Moreover, using a dataset is a passive action which cannot be used when testing active or autonomous methods. For these reasons, although (non-)rigid moving objects are common in real life, a lot of research in robotics still assumes a (semi-)static world. This greatly hinders efforts towards various research topics such as SLAM and navigation in dynamic environments, target tracking, and visual robot learning, thus limiting autonomy in robotics.

For these reasons, we developed a framework for Generating Realistic Animated Dynamic Environments --- GRADE~\cite{GRADE}. GRADE is a flexible, fully controllable, customizable, photorealistic, ROS-integrated framework to simulate and advance robotics research.

To demonstrate the limitations of current state-of-the-art dynamic SLAM methods, we used GRADE to: i) generate an \textbf{indoor} dynamic environment dataset by using only freely available assets and FUEL~\cite{zhou2021fuel}, ii) test popular indoor dynamic SLAM algorithms with some generated sequences to evaluate those and benchmark our work. While testing static sequences demonstrates that the data is usable by said frameworks, evaluating the dynamic ones proves that these methods cannot generalize to data different from the currently used benchmark datasets. We then experiment with different trained models of YOLOv5~\cite{yolov5} and Mask R-CNN~\cite{maskrcnn} and show that not always the best-performing one in terms of precision corresponds to the best ATE result. All of this while focusing on a metric that is often overlooked by the community: the amount of time the SLAM framework is capable of tracking the trajectory, which is an essential indication of the robustness of the considered method and helps to create a contextualization of the results.
	\section{Related Work}
\label{sec:soa}

Historically, one of the core robotics problems is mapping an unknown environment. A lot of the current SLAM research still focuses on static environments~\cite{surveySlam}, despite the belief of this being a solved problem, and how to actively explore them~\cite{placed2022survey}. Lately, visual SLAM has gained traction with respect to other methods, with RTABMap~\cite{labbe2019rtab} and ORB-SLAM~\cite{murORB2} which are just two among all the possible frameworks that can be used to perform it. Most of the current perception-based methods are developed in static environments and are expected to fail or degrade in dynamic ones, making them hard to be used in real-world everyday scenarios. Indeed, tracking the camera trajectory of the robot in dynamic environments is a notoriously difficult problem~\cite{surveyVodom}. Nonetheless, research in SLAM for dynamic worlds has still limited (although increasing) traction, mainly due to difficulties in simulating data and the inherent danger of directly testing an autonomous method in the real world. Many methods addressing dynamic worlds rely on segmentation or optical clues to filter out features of dynamic subjects, and most have no real-time capabilities. Among those, one of the most successful is DynaSLAM~\cite{dynaslam}, which uses Mask R-CNN~\cite{maskrcnn} and multi-view geometry. DynamicVINS~\cite{dynamic-vins} employs YOLOv5~\cite{yolov5} to mask the features belonging to dynamic objects. StaticFusion~\cite{staticfusion} instead relies on pointclouds clustering segmentation to work. Learning-based methods, such as TartanVO~\cite{tartanvo}, propose to learn visual odometry on synthetic and real data to reconstruct the robot trajectory. However, the limited availability of testing sequences and environments makes those fail when applied to different situations or environments, as shown also in~\cite{tartanvo}.
	\section{Own approach and contributions}
\label{sec:exp}
As explained thoroughly in GRADE~\cite{GRADE}, we used our framework to generate a dataset of indoor dynamic sequences autonomously recorded using FUEL~\cite{zhou2021fuel}. With these evaluations, we want to demonstrate that we can use the data generated in robotic applications by testing state-of-the-art dynamic SLAM methods, and highlight the current limitations of such frameworks. 
We selected two static SLAM methods, RTABMap~\cite{labbe2019rtab} and ORB-SLAMv2~\cite{murORB2}, to demonstrate that the visual information is not misleading by itself when testing static sequences and that the data is usable for the visual odometry task. Then we picked DynaSLAM, which uses Mask R-CNN to segment dynamic content, DynamicVINS, which instead uses YOLO, StaticFusion, i.e. a non-learning based method that performs clustering on the pointclouds, and TartanVO which, although it is not a proper SLAM system, is a learned visual odometry method developed specifically for challenging scenarios. DynamicVINS was tested in both its VO and VIO variations and with a minor modification to account for possible failures~\cite{GRADE}.

We used the generated data to train both YOLOv5 and Mask R-CNN, the networks used in~\cite{dynamic-vins} and~\cite{dynaslam}. We used the synthetic data both to train them from scratch and as pre-training step. Using the resulting network weights we evaluate the corresponding SLAM method, i.e. DynaSLAM with Mask R-CNN and Dynamic VINS with YOLOv5, on \textit{fr3/walking} sequences showing contrasting results.

\subsection{SLAM}
\label{sec:evalSLAM}

We select four RGBD sequences from the GRADE dataset~\cite{GRADE}, in which the robot stays horizontal (H) and four in which the robot is free to move. Each sequence are 60 seconds long marked as static (S), dynamic without flying objects (D), with flying objects (F) and with occlusions of the camera (WO). We perform evaluations of both groundtruth data and with added noise. Depth data was limited both to $3.5$ meters, which is a reasonable value when using for example a RealSense D435i, and $5$ meters. The added noise to the depth values is based on the model described in~\cite{intelnoisemodel}. To the RGB data we add random rolling shutter noise ($\mu=0.015$, $\sigma=0.006$), and blur following~\cite{deblurDatasetIMU}. The IMU drift and noise parameters are taken from~\cite{rotors}. Image data was recorded at 30 Hz, IMU at 240 Hz, and groundtruth pose at 60 Hz. As evaluation metrics, we utilize the ATE RMSE and the amount of time the framework can successfully track the trajectory. The latter is a critical evaluation quantity to be considered. It helps the reader put ATE values in perspective whenever the framework fails due to some featureless frames or occlusions. For consistency, when evaluating DynamicVINS, we considered only experiments in which the first initialization was successful. We first analyze the results with the depth limited at $3.5$ meters. We report them on Tab.~\ref{tab:slamgt} and Tab.~\ref{tab:slamnoisy}. One can notice that all the methods perform poorly in the majority of the sequences. Focusing on \textit{noisy} experiments, which are more related to reality, we can see that with static sequences most of the methods perform well, except TartanVO and StaticFusion which fail. Furthermore, one should not be misled by the good ATE results of DynaSLAM on dynamic sequences. Indeed, in 5 out of 8 experiments the camera lost track of the trajectory for at least $\sim27$ seconds (three times more than $49$ s) and performing sometimes worse than the ORB-SLAMv2, i.e. its underlying mapping framework. In general, we can infer that, despite these methods showing compelling results when tested with other datasets, they exhibit several limitations when tested on different data. The fact that the methods perform well with the static sequences demonstrates how it is not a problem of the data used, but it is a problem inherent to the dynamic nature of the environment or the presence of featureless frames. Overall, DynamicVINS seems to be the best-performing method when considering both ATE and the time missing from each experiment. However, despite the help of the IMU, in DH sequence the ATE is over $1.6$ meters for just a 60 s sequence. By comparing the tests performed on groundtruth and noisy data one can see that in the majority of the experiments the noisy ones perform slightly worse. However, in general, the results are similar and one can draw the same overall conclusions. Finally, we can compare corresponding sequences with the depth limited to 3.5 and 5 meters using tables Tab.~\ref{tab:slamgt5m} and Tab.~\ref{tab:slamgt} for the ground truth data, and Tab.~\ref{tab:slamnoisy5m} and Tab.~\ref{tab:slamnoisy5m} for the noisy one. As expected, TartanVO yields equal results, by being a method that works only on visual data. RTABMap shows performance which are greatly degraded in all sequences, in both missing time and ATE. ORB-SLAM and DynaSLAM(VIO) are, for the most part, comparable. We can also notice how DynaSLAM(VO) shows worse performance, indicating the reliance of the VIO counterpart on the IMU. StaticFusion, as shown also in other works like~\cite{runz2018maskfusion}, shows degrading performance with increased depth data. DynaSLAM seems overall the most stable, except for the D-noisy sequence which shows high variability.

\subsection{Network models variations}
We will consider here four models: S-COCO, S-GRADE, S-GRADE+S-COCO and S-GRADE+COCO. These correspond to different training strategies with a reduced coco dataset (S-COCO), a reduced GRADE dataset (S-GRADE), COCO, and combination of pretraining and finetuning (S-GRADE+[S-COCO, COCO]). We refer the reader to~\cite{GRADE} for additional insights. In general, the best performing models for both YOLO and Mask RCNN when tested on the TUM RGBD labelled data, were obtained with S-GRADE+COCO, followed by COCO (BASELINE in~\cite{GRADE}), S-GRADE+S-COCO, S-GRADE, S-COCO.

\subsubsection{YOLOv5 and Dynamic VINS}
We utilize them with Dynamic VINS to evaluate their performance with the TUM \textit{fr3/walking} sequences. The results, presented in Tab.~\ref{tab:dyna-yolo}, are averaged among three runs. The baseline values are not the ones from~\cite{dynamic-vins} since we were unable to reproduce them for the \textit{rpy} and \textit{static} sequences. One can easily notice how the models pre-trained on the synthetic data consistently obtain results which are at par or better than the baseline model. Surprisingly, using the model trained on S-GRADE, despite showing the lowest detection performance among the models considered in this test, is the best performing one in two sequences, remarking the fact that more research is necessary on dynamic SLAM. In this case, between the tests, there was no difference in the percentage amount of the successfully tracked trajectory.

\begin{table}[h]
    \centering
    \resizebox{\columnwidth}{!}{
\begin{tabular}{l|c|c|c|c|c|c}
            Model & S-COCO & S-GRADE & \begin{tabular}{c}S-GRADE\\+ S-COCO\end{tabular} & \begin{tabular}{c}S-GRADE\\+ COCO\end{tabular} & Baseline & $\%$ Traj. \\ \hline
 \textit{w.\_half} & 0.064 & \textbf{0.048} & \textit{0.059} & 0.066 & 0.069 & 87.81 \\
 \textit{w.\_xyz} & 0.052 & 0.050 & 0.049 & \textit{0.045} & \textbf{0.037} & 89.37 \\
 \textit{w.\_rpy} & 0.120 & 0.224 & \textit{0.116} & \textit{0.116} & \textbf{0.114} & 87.11\\
 \textit{w.\_stat} & 0.302 & \textbf{0.199} & 0.216 & \textit{0.203} & 0.218 & 90.18 \\
\end{tabular}}
\caption{ATE [m] and percentage of the tracked trajectory percentage of Dynamic VINS using our models on the \textit{fr3/walking} sequences. In \textbf{bold} the best one, in \textit{italics} the second best.}
    \label{tab:dyna-yolo}
\end{table}

\subsubsection{Mask R-CNN and DynaSLAM} 
Also in this case, we evaluate the performance of the trained model with the TUM \textit{fr3/walking} sequences using DynaSLAM. Each result is the average between three experiments, and all are shown in Tab.~\ref{tab:dyna-mrcnn}. Also in this case, we computed the baseline again, obtaining results which are close to the one reported in~\cite{dynaslam} with the exception of the \textit{walking\_rpy} sequence. However, this was necessary for us to be able to report the percentage amount of the tracked trajectory.  We used the trained models that showed the best performance in the segmentation task. Here, thanks to the offline nature of the method and the optimization procedure employed, ATE errors are much lower than the ones obtained from Dynamic VINS (see Tab.~\ref{tab:dyna-yolo}). By analysing the results and taking into consideration both the ATE and the amount of the tracked trajectory, one can see that in all cases there is a benefit to using a detection network which was pre-trained on synthetic data. However, a clear pattern cannot be derived yet, despite both S-GRADE + S-COCO and S-GRADE + COCO showing compelling results.

\begin{table}[h]
    \centering
    \resizebox{\columnwidth}{!}{
\begin{tabular}{l|c|c|c|c|c}
            & S-COCO & S-GRADE & \begin{tabular}{c} S-GRADE \\ + S-COCO \end{tabular} & \begin{tabular}{c} S-GRADE \\ + COCO \end{tabular} & Baseline \\ \hline
 \multirow{2}{*}{\textit{w.\_half}} & 0.031 & 0.034 & 0.032 & \textbf{0.028} & \textit{0.030} \\
 & 79.24 & \textit{89.37} & 78.65 & 89.34 & \textbf{89.53} \\ \hline
 \multirow{2}{*}{\textit{w.\_xyz}} & \textit{0.017} & \textit{0.017} & \textbf{0.016} & \textbf{0.016} & \textbf{0.016} \\
 & \textit{91.43} & 86.35 & \textbf{91.85} & 83.93 & 83.90 \\\hline
 \multirow{2}{*}{\textit{w.\_rpy}} & \textbf{0.034} & 0.104 & 0.060 & \textit{0.037} & 0.040 \\
 & 72.99 & \textbf{80.25} & 67.82 & \textit{77.79} & 75.68 \\\hline
 \multirow{2}{*}{\textit{w.\_stat}} & 0.010 & \textbf{0.007} & \textbf{0.007} & \textit{0.008} & \textbf{0.007} \\
 & \textbf{91.89} & \textit{91.61} & \textbf{91.89} & 77.77 & 89.06 \\
\end{tabular}}
\caption{ATE [m] and percentage of the tracked trajectory of DynaSLAM using our models on the \textit{fr3/walking} sequences. In \textbf{bold} the best, in \textit{italics} the second best.}
    \label{tab:dyna-mrcnn}
\end{table}

\begin{table*}[ht]
    \centering
    \resizebox{\textwidth}{!}{
\begin{tabular}{l|cc|cc|cc|cc|cc|cc|cc}
  & \multicolumn{2}{c|}{\begin{tabular}{@{}c@{}}Dynamic \\ VINS (VIO) \end{tabular}} & \multicolumn{2}{c|}{\begin{tabular}{@{}c@{}}Dynamic \\ VINS (VO) \end{tabular}} & \multicolumn{2}{c|}{TartanVO} & \multicolumn{2}{c|}{StaticFusion} & \multicolumn{2}{c|}{DynaSLAM} & \multicolumn{2}{c|}{ORB-SLAMv2}& \multicolumn{2}{c}{RTABMap} \\
Sequence & ATE [m] & \begin{tabular}{@{}c@{}}Missing \\ Time [s]\end{tabular} & ATE [m] & \begin{tabular}{@{}c@{}}Missing \\ Time [s]\end{tabular} & ATE [m] & \begin{tabular}{@{}c@{}}Missing \\ Time [s]\end{tabular} & ATE [m] & \begin{tabular}{@{}c@{}}Missing \\ Time [s]\end{tabular} & ATE [m] & \begin{tabular}{@{}c@{}}Missing \\ Time [s]\end{tabular} & ATE [m] & \begin{tabular}{@{}c@{}}Missing \\ Time [s]\end{tabular} & ATE [m] & \begin{tabular}{@{}c@{}}Missing \\ Time [s]\end{tabular} \\ \hline
FH  & 0.155 & 1.03 & 0.367 & 0.43 & 0.582 & 0.00 & 0.854 & 0.00 & 0.309 & 1.50 & 0.386 & 0.00 & 0.097 & 1.97\\
F  & 0.886 & 1.27 & 1.857 & 1.37 & 4.223 & 0.00 & 3.992 & 0.00 & 0.178 & 49.67 & 0.144 & 44.30 & 0.125 & 47.03\\ 
DH  & 1.681 & 0.43 & 1.183 & 3.30 & 1.234 & 0.00 & 1.091 & 0.00 & 0.002 & 57.33 & 0.005 & 56.80 & 0.013 & 49.13\\ 
D  & 0.707 & 0.67 & 1.598 & 1.63 & 1.356 & 0.00 & 2.278 & 0.00 & 0.043 & 26.93 & 0.700 & 11.07 & 0.405 & 28.77\\
WOH & 0.491 & 1.03 & 0.871 & 1.23 & 2.399 & 0.00 & 1.826 & 0.00 & 0.023 & 28.53 & 0.022 & 27.90 & 0.101 & 27.93 \\
WO  & 1.086 & 2.63 & 1.163 & 3.17 & 2.473 & 0.00 & 2.213 & 0.00 & 0.119 & 55.23 & 0.171 & 48.23 & 0.075 & 50.70\\
SH  & 0.419 & 0.57 & 0.069 & 0.43  & 2.517 & 0.00 & 4.184 & 0.00 & 0.016 & 0.00 & 0.018 & 0.00 & 0.072 & 0.00\\ 
S  & 0.177 & 0.57 & 0.137 & 0.43  & 1.308 & 0.00 & 3.538 & 0.00 & 0.029 & 0.00 & 0.026 & 0.00 & 0.133 & 0.00\\
\end{tabular}}
    \caption{ATE RMSE [m] and missing time [s] of the tested \textbf{noisy} sequences. Each experiment is 60 seconds long. Depth limited to 3.5 m. \bigskip}
    \label{tab:slamnoisy}
\end{table*}

\begin{table*}[ht]
    \centering
    \resizebox{\textwidth}{!}{
\begin{tabular}{l|cc|cc|cc|cc|cc|cc|cc}
  & \multicolumn{2}{c|}{\begin{tabular}{@{}c@{}}Dynamic \\ VINS (VIO) \end{tabular}} & \multicolumn{2}{c|}{\begin{tabular}{@{}c@{}}Dynamic \\ VINS (VO) \end{tabular}} & \multicolumn{2}{c|}{TartanVO} & \multicolumn{2}{c|}{StaticFusion} & \multicolumn{2}{c|}{DynaSLAM} & \multicolumn{2}{c|}{ORB-SLAMv2}& \multicolumn{2}{c}{RTABMap} \\
Sequence & ATE [m] & \begin{tabular}{@{}c@{}}Missing \\ Time [s]\end{tabular} & ATE [m] & \begin{tabular}{@{}c@{}}Missing \\ Time [s]\end{tabular} & ATE [m] & \begin{tabular}{@{}c@{}}Missing \\ Time [s]\end{tabular} & ATE [m] & \begin{tabular}{@{}c@{}}Missing \\ Time [s]\end{tabular} & ATE [m] & \begin{tabular}{@{}c@{}}Missing \\ Time [s]\end{tabular} & ATE [m] & \begin{tabular}{@{}c@{}}Missing \\ Time [s]\end{tabular} & ATE [m] & \begin{tabular}{@{}c@{}}Missing \\ Time [s]\end{tabular} \\ \hline
FH  & 0.069 & 0.67 & 0.201 & 0.43 & 0.551 & 0.00 & 0.085 & 0.00 & 0.241 & 3.43 & 0.149 & 0.00 & 0.126 &  0.00 \\ 
F   & 0.647 & 1.20 & 1.337 & 2.93 & 4.132 & 0.00 & 2.866 & 0.00 & 0.147 & 49.23 & 0.431 & 39.10 & 0.115 &  46.93\\ 
DH  & 8.103 & 0.43 & 1.178 & 8.17 & 1.259 & 0.00 & 1.664 & 0.00 & 0.008 & 57.07 & 0.005 & 48.53 & 0.094 &  21.63 \\ 
D   & 0.188 & 0.67 & 1.304 & 0.83 & 1.264 & 0.00 & 1.212 & 0.00 & 0.057 & 6.33 & 0.459 & 0.30 & 0.492 &  7.07 \\
WOH & 0.239 & 1.10 & 1.272 & 1.00 & 2.361 & 0.00 & 1.980 & 0.00 & 0.015 & 27.70 & 0.012 & 27.73 & 0.042 &  25.87 \\
WO  & 0.501 & 2.20 & 0.985 & 3.47 & 2.380 & 0.00 & 2.807 & 0.00 & 0.083 & 55.20 & 0.163 & 48.23 & 0.053 &  50.60 \\
SH  & 0.109 & 0.57 & 0.023 & 0.43 & 2.395 & 0.00 & 0.594 & 0.00 & 0.016 & 0.00 & 0.012 & 0.00 & 0.039 &  0.13 \\ 
S   & 0.205 & 0.57 & 0.039 & 0.43 & 1.205 & 0.00 & 7.919 & 0.00 & 0.010 & 0.00 & 0.011 & 0.00 & 0.043 &  0.00 \\
\end{tabular}}
    \caption{ATE RMSE [m] and missing time [s] of the tested sequences \textbf{w/o added noise}. Each experiment is 60 seconds long. Depth limited to 3.5 m.}
    \label{tab:slamgt}
\end{table*}

\begin{table*}[ht]
    \centering
    \resizebox{\textwidth}{!}{
\begin{tabular}{l|cc|cc|cc|cc|cc|cc|cc}
  & \multicolumn{2}{c|}{\begin{tabular}{@{}c@{}}Dynamic \\ VINS (VIO) \end{tabular}} & \multicolumn{2}{c|}{\begin{tabular}{@{}c@{}}Dynamic \\ VINS (VO) \end{tabular}} & \multicolumn{2}{c|}{TartanVO} & \multicolumn{2}{c|}{StaticFusion} & \multicolumn{2}{c|}{DynaSLAM} & \multicolumn{2}{c|}{ORB-SLAMv2}& \multicolumn{2}{c}{RTABMap} \\
Sequence & ATE [m] & \begin{tabular}{@{}c@{}}Missing \\ Time [s]\end{tabular} & ATE [m] & \begin{tabular}{@{}c@{}}Missing \\ Time [s]\end{tabular} & ATE [m] & \begin{tabular}{@{}c@{}}Missing \\ Time [s]\end{tabular} & ATE [m] & \begin{tabular}{@{}c@{}}Missing \\ Time [s]\end{tabular} & ATE [m] & \begin{tabular}{@{}c@{}}Missing \\ Time [s]\end{tabular} & ATE [m] & \begin{tabular}{@{}c@{}}Missing \\ Time [s]\end{tabular} & ATE [m] & \begin{tabular}{@{}c@{}}Missing \\ Time [s]\end{tabular} \\ \hline
FH  & 0.179 & 0.67 & 0.349 & 0.43 & 0.568 & 0.00 & 0.059 & 0.00  & 0.200 & 4.50  & 0.291 & 0.00 & 0.086 & 15.07 \\
F   & 0.904 & 1.63 & 2.315 & 2.13 & 4.192 & 0.00 & 2.781 & 0.00  & 0.189 & 45.53 & 0.129 & 43.00 & 0.125 & 50.63 \\ 
DH  & 1.749 & 0.43 & 2.047 & 3.40 & 1.214 & 0.00 & 14.938 & 0.00 & 0.002 & 57.40 & 0.005 & 56.73 & 0.030 & 50.07 \\ 
D   & 0.611 & 0.67 & 1.616 & 1.23 & 1.350 & 0.00 & 22.374 & 0.00 & 0.109 & 5.97  & 0.652 & 5.13 & 0.171 & 41.73 \\
WOH & 0.561 & 1.27 & 1.550 & 0.90 & 2.389 & 0.00 & 4.926 & 0.00  & 0.025 & 27.83 & 0.022 & 27.87 & 0.047 & 33.30 \\
WO  & 0.962 & 2.40 & 1.429 & 2.40 & 2.399 & 0.00 & 1.418 & 0.00  & 0.112 & 55.23 & 0.142 & 48.20 & 0.041 & 52.53 \\
SH  & 0.404 & 0.57 & 0.063 & 0.43 & 2.537 & 0.00 & 2.721 & 0.00  & 0.017 & 0.00  & 0.017 & 0.00 & 0.061 & 17.13 \\ 
S   & 0.199 & 0.57 & 0.066 & 0.43 & 1.259 & 0.00 & 22.282 & 0.00 & 0.029 & 0.00  & 0.027 & 0.00 & 0.220 & 11.53 \\
\end{tabular}}
    \caption{ATE RMSE [m] and missing time [s] of the tested \textbf{noisy} sequences. Each experiment is 60 seconds long.  Depth limited to 5 m. \bigskip}
    \label{tab:slamnoisy5m}
\end{table*}

\begin{table*}[ht]
    \centering
    \resizebox{\textwidth}{!}{
\begin{tabular}{l|cc|cc|cc|cc|cc|cc|cc}
  & \multicolumn{2}{c|}{\begin{tabular}{@{}c@{}}Dynamic \\ VINS (VIO) \end{tabular}} & \multicolumn{2}{c|}{\begin{tabular}{@{}c@{}}Dynamic \\ VINS (VO) \end{tabular}} & \multicolumn{2}{c|}{TartanVO} & \multicolumn{2}{c|}{StaticFusion} & \multicolumn{2}{c|}{DynaSLAM} & \multicolumn{2}{c|}{ORB-SLAMv2}& \multicolumn{2}{c}{RTABMap} \\
Sequence & ATE [m] & \begin{tabular}{@{}c@{}}Missing \\ Time [s]\end{tabular} & ATE [m] & \begin{tabular}{@{}c@{}}Missing \\ Time [s]\end{tabular} & ATE [m] & \begin{tabular}{@{}c@{}}Missing \\ Time [s]\end{tabular} & ATE [m] & \begin{tabular}{@{}c@{}}Missing \\ Time [s]\end{tabular} & ATE [m] & \begin{tabular}{@{}c@{}}Missing \\ Time [s]\end{tabular} & ATE [m] & \begin{tabular}{@{}c@{}}Missing \\ Time [s]\end{tabular} & ATE [m] & \begin{tabular}{@{}c@{}}Missing \\ Time [s]\end{tabular} \\ \hline
FH  & 0.073 & 0.67 & 0.259 & 0.43 & 0.551 & 0.00 & 0.059 & 0.00 &  0.221 & 0.30   & 0.199 & 0.00 & 0.229 &  7.53 \\ 
F   & 1.814 & 1.40 & 1.362 & 1.77 & 4.132 & 0.00 & 2.781 & 0.00 &  0.228 & 45.10  & 0.512 & 37.13 & 0.129 &  48.33 \\ 
DH  & 7.492 & 0.43 & 1.811 & 7.50 & 1.259 & 0.00 & 14.938 & 0.00 & 0.008 & 54.40  & 0.009 & 49.53 & 0.108 &  48.33 \\ 
D   & 0.201 & 0.67 & 0.738 & 0.43 & 1.264 & 0.00 & 22.374 & 0.07 & 0.038 & 4.33  & 0.275 & 0.53 & 0.154 &  23.40 \\
WOH & 0.228 & 1.10 & 1.256 & 2.57 & 2.361 & 0.00 & 4.926 & 0.00 &  0.012 & 27.70  & 0.015 & 27.70 & 0.053 &  30.70 \\
WO  & 0.679 & 2.43 & 1.031 & 3.10 & 2.473 & 0.00 & 1.418 & 0.00 &  0.107 & 55.10  & 0.138 & 48.20 & 0.025 &  51.53 \\
SH  & 0.121 & 0.57 & 0.023 & 0.43 & 2.395 & 0.00 & 2.721 & 0.07 &  0.012 & 0.00   & 0.011 & 0.00 & 0.019 &  10.60 \\ 
S   & 0.226 & 0.57 & 0.035 & 0.43 & 1.205 & 0.00 & 22.282 & 0.00 & 0.011 & 0.00   & 0.014 & 0.00 & 0.018 &  12.60 \\
\end{tabular}}
    \caption{ATE RMSE [m] and missing time [s] of the tested sequences \textbf{w/o added noise}. Each experiment is 60 seconds long. Depth limited to 5 m.}
    \label{tab:slamgt5m}
\end{table*}
	
\section{CONCLUSIONS}
\label{sec:conc}
In this work, by using data from~\cite{GRADE}, we have shown how all the tested models methods fail, one way or another, to successfully generalize to our dynamic sequences. However, the majority of them are capable of tracking static trajectories, signifying that the issue is not on the realism or the quality of the data itself but lies possibly on the parameter tuning and, more in general, on the method themselves. We have also shown that the models trained on our synthetic data can bring a performance improvement when a model trained with that is used with the corresponding SLAM framework. However, this also indicates that more research in this regard is needed since there are instances in which networks trained \textit{only} on either S-COCO and S-GRADE perform the best, despite these being the worst-performing models in the detection and segmentation tasks.

\end{document}